\documentclass[10pt,twocolumn,letterpaper]{article}

\usepackage{iccv}
\usepackage{times}
\usepackage{epsfig}
\usepackage{graphicx}
\usepackage{amsmath}
\usepackage{amssymb}
\usepackage{multirow}
\usepackage{makecell}
\usepackage{booktabs}
\usepackage[ruled]{algorithm2e}
\usepackage{algpseudocode}
\usepackage{makeidx}
\usepackage{balance}
\newcommand\SET{\mathcal}

\usepackage[pagebackref=true,breaklinks=true,letterpaper=true,colorlinks,bookmarks=false]{hyperref}

\iccvfinalcopy 

\def\iccvPaperID{6838} 

\ificcvfinal\fi
\makeindex
\begin{document}

\title{Global Correlation Network:\\
End-to-End Joint Multi-Object Detection and Tracking}

\author{Xuewu Lin$^1$, Yu-ang Guo$^2$, Jianqiang Wang$^1$*\\
$^1$Intell Tsinghua University, $^2$ Beihang University\\
Beijing Haidian, China\\
{*\tt\small wjqlws@tsinghua.edu.cn}
}

\maketitle
\ificcvfinal\thispagestyle{empty}\fi

\begin{abstract}
Multi-object tracking (MOT) has made great progress in recent years, 
but there are still some problems. Most MOT algorithms follow tracking-by-detection
framework, which separates detection and tracking into two independent parts. 
Early tracking-by-detection algorithms need to do two feature extractions for detection 
and tracking. Recently, some algorithms make the feature extraction into one network,
but the tracking part still relies on data association and needs complex post-processing
for life cycle management. Those methods do not combine detection and tracking well. 
In this paper, we present a novel network to realize joint multi-object detection and 
tracking in an end-to-end way, called Global Correlation Network (GCNet).
Different from most object detection methods, GCNet introduces the global correlation layer for regression of absolute size and coordinates of bounding boxes instead of offsets prediction.
The pipeline of detection and tracking by GCNet is conceptually simple, which does not need
non-maximum suppression, data association, and other complicated tracking strategies. GCNet was
evaluated on a multi-vehicle tracking dataset, UA-DETRAC\cite{wen_ua-detrac_2015}, and demonstrates promising
performance compared to the state-of-the-art detectors and trackers.
\end{abstract}
\begin{figure*}[hptb]
  \centering
  \includegraphics[scale=0.285]{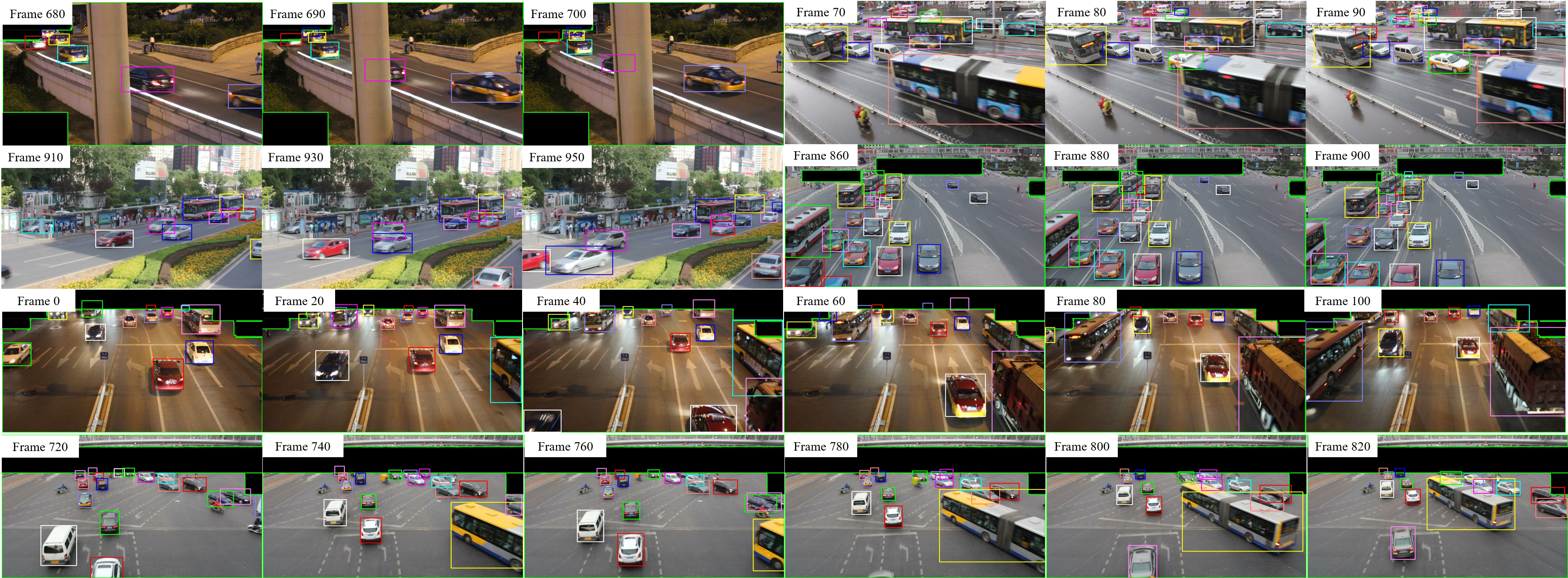}
  \caption{Examples of tracking results on UA-DETRAC dataset. Eight different secenarios are shown here, and each scene shows three frames of images. The bounding boxes with different colors represent different vehicle IDs, and the black areas indicate areas that are ignored defined in UA-DETRAC.}
  \label{fig:trackingresults}
\end{figure*}
\section{Introduction}
MOT is a basic problem in computer vision, and its goal is to calculate the trajectories of all the interested objects from the consecutive frames of images. It has a wide range of application scenarios, such as autonomous driving, motion attitude analysis and traffic monitoring, and it receives increasing attention in recent years.

Traditional MOT algorithms follow tracking-by-detection framework, which is split into two modules, detection and tracking. With the development of object detection, those algorithms achieve great performance and dominate almost the entire MOT domain. The tracking module in tracking-by-detection framework usually contains three parts: feature extraction, data association and lifecycle management. Early tracking methods use simple features to accomplish data association, such as location, shape and velocity, but those features have obvious deficiencies. Later, most methods utilize appearance features, especially high-level features from deep neural networks. Those appearance features can greatly improve association accuracy and robustness, but it leads to increased calculation. Currently, some MOT algorithms integrate feature extraction into the detection module, which add the ReID head to get the instance-level feature for data association. Although those algorithms require less computation, the data association still needs to do motion prediction and set complex tracking strategies, resulting in too many hyperparameters and a cumbersome inference pipeline.

In this paper, we present a novel network for end-to-end joint detection and tracking. The network realizes bounding box regression and tracking in the same way, which is called global correlation. It is we all know that bounding box regress usually use the local feature to estimate offsets between the anchor and the ground truth, or estimate box size and the offset between key point and feature location. In our framework, we want to regression absolute coordinate and size of bounding box rather than relative coordinate or offset. But in traditional convolutional neural networks, the local feature cannot contain global information when the receptive field is not large enough. The self-attention mechanism allows the features of each location to contain global information, but its computational complexity is too large to be used on a high-resolution feature map. Here, we introduced the global correlation layer to encode global information into features at each location. For a feature map $F\in R^{h\times w \times c}$, two feature maps Q and K are obtained after two linear transformations.
\begin{equation}
    Q_{ij}=W_{q}F_{ij}, K_{ij}=W_{k}F_{ij}
\end{equation}
where $X_{ij}\in R^{c}$, denotes the feature vector at the $i_{th}$ row and the $j_{th}$ column of $X$. Then for each feature vector $Q_{ij}$, calculate the cosine distance between it and all $K_{ij}$, after another linear transformation $W$, we can get the correlation vectors $C_{ij}\in R^{c'}$.
\begin{equation}\label{eq:correlation}
C_{ij}=W\cdot \text{flatten}\left(
\begin{bmatrix}
\frac{Q_{ij} K_{11}}{|Q_{ij}| |K_{11}|}  & \dots & \frac{Q_{ij} K_{1w}}{|Q_{ij}| |K_{1w}|} \\
\vdots & \ddots & \vdots \\
\frac{Q_{ij} K_{h1}}{|Q_{ij}| |K_{h1}|}  & \dots & \frac{Q_{ij} K_{hw}}{|Q_{ij}| |K_{hw}|} 
\end{bmatrix}\right)
\end{equation}

Those correlation vectors $C_{ij}$ have been encoded the correlation between local feature vectors $Q_{ij}$ with global feature map $K$, so it can be used to regress the absolute bounding boxes for the objects at the corresponding positions in the image. All of the correlation vector $C_{ij}$ can form a correlation map $C\in R^{h\times w\times c'}$, so we can simply get bounding boxes $B\in R^{h\times w\times 4}$ by a convolution layer with $1\times 1$ kernel size. When performing object detection, the $K$ and $Q$ from the image in the same frame will be used; when performing object tracking, the $Q$ from the image in the previous frame and $K$ from the image in the current frame will be used. In this way, we unify detection and tracking under the same framework.

In terms of object classification brunch, we use the same network structure and training strategy as CenterNet. When infers, we will get a detection heatmap $Y_d$ and a tracking heatmap $Y_t$ in each frame. The detection heatmap $Y_{d}$ denotes detection confidences of object centers in the current frame, and the tracking heatmap $Y_t$ denotes tracking confidence between the current frame and the next frame. Peaks in the heatmaps correspond to detection and tracking key points, and the max-pooling will be used to get final bounding boxes without box non-maximum suppression (NMS).
\begin{equation}
B_{ij} \in Result \quad \forall i,j \rightarrow maxpool\left(Y,3,1\right)_{ij}=Y_{ij}
\end{equation}
where the $maxpool(H,a,b)$ represents a max-pooling layer with kernel size $a$ and stride $b$. So, the GCNet can realize joint multi-object detection and tracking without complicated post-process, such as NMS and data association, which has a concise pipeline.

We started from the perspective of autonomous driving, so we performed algorithm evaluation on a vehicle tracking dataset, UA-DETRAC. And GCNet shows competitive performance with $74.04\%$ AP and $36$ FPS in detection, $19.10\%$ PR-MOTA and $34$ FPS in tracking. Figure \ref{fig:trackingresults} shows some examples of tracking results.
\begin{figure*}[t]
  \centering
  \includegraphics[scale=0.5]{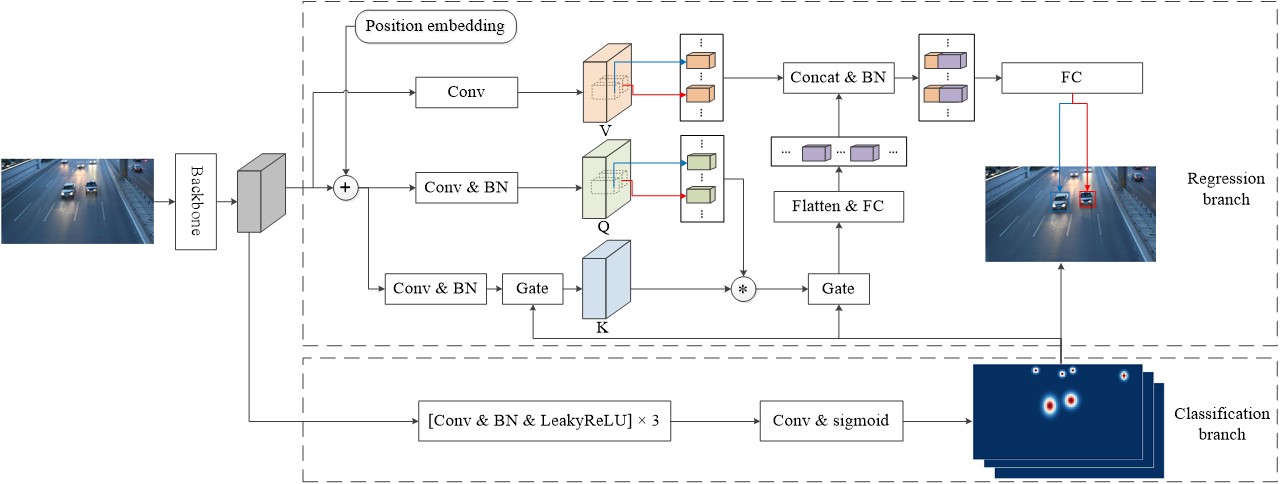}
  \caption{Detection Module Architecture. The Conv block denotes a convolution layer, $BN$ denotes batch normalization, $FC$ is a full-connected layer. This module takes an image $I\in R^{h\times w\times 3}$ as inputs, outputs a center detection confidence map $Y_d\in R^{h' \times w'\times n}$ ($n$ is the number of categories) and a bunch of bounding boxes $B_d\in R^{h' \times w'\times 4}$. When training, it produces all of the potential bounding boxes in a tensor $B_d\in R^{h'\times w'\times 4}$. And at test time, it just calculates the bounding boxes at peaks of the confidence map.}
  \label{fig:detectionmodule}
\end{figure*}
\section{Related Works}
Object detection: With the explosion of deep learning, object detection technology has developed rapidly. Existing object detection algorithms can be divided into two categories, anchor-based algorithms \cite{ren_faster_2017}\cite{lin_focal_2017}\cite{girshick_fast_2015}\cite{redmon_yolov3_2018}\cite{fu_dssd_2017}\cite{he_k__gkioxari_g__piotr_dollar_et_al_mask_2017} and anchor-free algorithms \cite{zhou_objects_2019}\cite{law_cornernet_2018}\cite{carion_end--end_2020}. Anchor-based algorithms set a series of anchor boxes and regress offsets between anchor boxes and ground truth by local features. Most anchor-free algorithms use full-convolution networks to estimate the key points of targets, and then obtain the bounding boxes through the key points. Those algorithms consider local features for bounding box regression, so they can only get the offsets between anchor boxes or key points and ground truth rather than absolution bounding box coordination. DETR \cite{carion_end--end_2020} adopt an encoder-decoder architecture based on transformers to achieve object detection. Transformer can integrate the global information into the features at each position, but the self-attention mechanism of transformer requires a lot of computation and GPU memory, which is difficult to apply to high-resolution feature maps. In our joint detection and tracking framework, we should not only detect objects in a single image, but also tracking objects in different images, and the offsets for the same object in different images are hard to define. So, here we introduce the global correlation layer to embed global information into features at each position for absolution coordinates regression, which can apply to higher-resolution feature maps than transformer.

Tracking-by-detection: With the improvement of detection accuracy, tracking-by-detection methods \cite{rezatofighi_joint_2015}\cite{bewley_simple_2016}\cite{wojke_simple_2017}\cite{bochinski_high-speed_2017}\cite{ullah_hog_2016}\cite{ristani_features_2018}\cite{shi_rank-1_2019} have become the mainstream in the field of multi-object tracking. Tracking is considered as a data association problem in the tracking-by-detection framework. Features such as motion, shapes and appearance are used to describe the correlation between detections and tracks and establish a correlation matrix. Hungary algorithm\cite{sadeghian_tracking_2017}, JPDA\cite{rezatofighi_joint_2015}, MHT\cite{kim_multiple_2015} and other algorithms take the correlation matrix as input to complete data association. Although those algorithms have made great progress, there are some drawbacks. First, they do not combine detector and tracker well and most of them need to do feature extraction separately, which brings unnecessary computation. Second, they often rely on complicated tracking rules for lifecycle management, resulting in numerous hyperparameters and difficult tuning. In our approach, detection and tracking are done in the same way, so they are well combined and the computation of feature extraction is reduced. And our approach gets rid of the complex tracking rules.

Joint detection and tracking: In the field of multi-object tracking, it is an important research direction to combine detection and tracking. \cite{zhang_fairmot_2020}\cite{lu_retinatrack_2020}\cite{voigtlaender_mots_2019}\cite{wang_towards_2020} start from multi-task learning, and add ReID feature extraction to the existing object detection networks. \cite{bergmann_tracking_2019} adopt Faster-RCNN framework, and accomplish tracking by RoI pooling and bounding box regression without data association. \cite{zhou_tracking_2020} takes the current frame, the previous frame, and a heatmap rendered from tracked object centers as inputs, and produces an offset map which greatly simplifies data association. \cite{peng_chained-tracker_2020} convert the MOT problem into pair-wise object detection problem, and realize end-to-end joint object detection and tracking. \cite{zhou_tracking_2020} and \cite{peng_chained-tracker_2020} are two innovative approaches to joint detection and tracking. Like them, this paper provides a new idea for joint detection and tracking.
\begin{figure*}[htbp]
  \centering
  \includegraphics[scale=0.25]{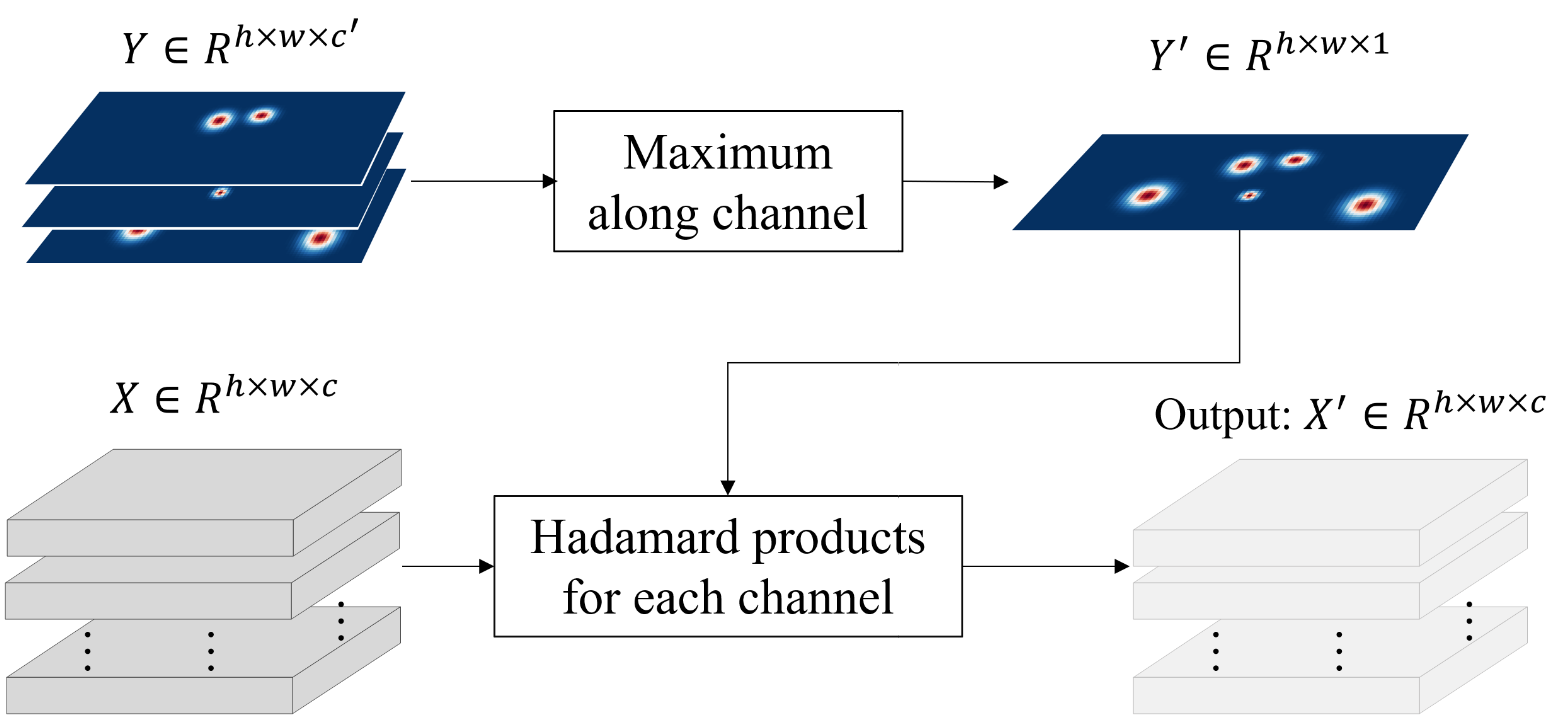}
  \caption{Illustration of Gate step. This step is similar with sSE block in \cite{roy_concurrent_2018}. The difference is the inputs of the Gate step are two tensors. Tensor $X$ contains the values and tensor $Y$ denotes spatial attention. In our detection module, the confidence map from classification branch can be directly regarded as spatial attention.}
  \label{fig:gate}
\end{figure*}
\section{Methodology}
Our network is designed to solve the online MOT problem. At time step $t$, we have obtained the object trajectories $\{T_1,T_2,…,T_n \}$ from time $0$ to time $t-1$, where $T_i=[B_{i,1},B_{i,2},\dots ,B_{i,t-1}]$ and $B_{ij}$ is the bounding box of object $i$ at time $j$.
Given an image of the current frame $I_t\in R^{h\times w\times 3}$, we should assign the bounding boxes $B_{x,t}$ of objects in the current frame to historical trajectories or generate new trajectories.
The following will introduce our algorithm in detail.
\subsection{Global Correlation Network}
\textbf{Detection module:} The detection module architecture is depicted as Figure \ref{fig:detectionmodule}, which contains three parts, backbone, classification branch and regression branch. Backbone is for high-level feature extraction. Because our classification is the same as CenterNet, each location of the feature map may correspond to an object center point, resolution of the feature map has a crucial impact on network performance. To get high resolution and keep large receptive field, we take the same skip connection structure as FPN, but only output the finest level feature map $F$. The size of the feature map $F$ is $h'\times w'\times c$, equal to $\frac{h}{8}\times \frac{w}{8}\times c$, where $h$ and $w$ are height and width of the original image. This resolution is $4$ times that of DETR. The classification branch is a full convolution network and outputs a confidence map $Y_d\in R^{h'\times w'\times n}$ with values between $0$ and $1$. The peaks of the $i_{th}$ channel of $Y_d$ correspond to the centers of objects belonging to the $i_{th}$ category. Regression branch is to calculate bounding boxes $\{[x,y,h,w]_i  | 1\leqslant i\leqslant N\}$. First, take $F$ and $Y_d$ as inputs, three feature maps $K$, $Q$, $V$ are generated.
\begin{equation} \label{eq:qkv}
\begin{split}
& Q = BN_Q(Conv_Q(F,1,1,c)+P) \\
& K = Gate[BN_K(Conv_K(F,1,1,c)+P),Y_d] \\
& V = Conv_V(F,1,1,c)
\end{split}
\end{equation}
where $Conv(F,a,b)$ denotes convolution layer with kernel size $a$, strides $b$ and kernel number $c$, $BN$ denotes batch normalization layer. $Gate(X,Y)$ is depicted as Figure \ref{fig:gate}, which is a kind of spatial attention. $P$ is the position embedding with same shape as $F$, and its expression is shown in equation (\ref{eq:position_embedding}).
\begin{equation}\label{eq:position_embedding}
\begin{split}
P_{ijk} = \begin{cases}
\cos \left(\frac{4\pi k}{c}+\frac{\pi i}{h}\right), & 0 \leqslant k < \frac{c}{2} \\
\cos \left(\frac{4\pi k}{c}+\frac{\pi j}{w}\right), & \frac{c}{2} \leqslant k < c
\end{cases} \\
0 \leqslant i < h', 0 \leqslant  j <w'
\end{split}
\end{equation}

Two embedding vectors that are close in position have a large cosine similarity, and two that are farther away have a small cosine similarity. This attribute can reduce the negative impact of similar objects when tracking. Then calculate the correlation vectors $C_{ij}$ between the $Q_{ij}$ and $K$ by equation (\ref{eq:correlation}). The final bounding boxes $B_{d,ij}=[x_{ij},y_{ij},h_{ij},w_{ij}]$ can be get by equation (\ref{eq:boundingbox_output}). Here, we directly regress the absolute coordinates and size of the bounding box, which is different from most existing methods, especially the anchor-based methods.
\begin{equation} \label{eq:boundingbox_output}
B_{d,ij} = W \cdot BN\left(\begin{bmatrix} C_{ij} & V_{ij}\end{bmatrix}\right)
\end{equation}

\begin{figure*}[t]
  \centering
  \includegraphics[scale=0.32]{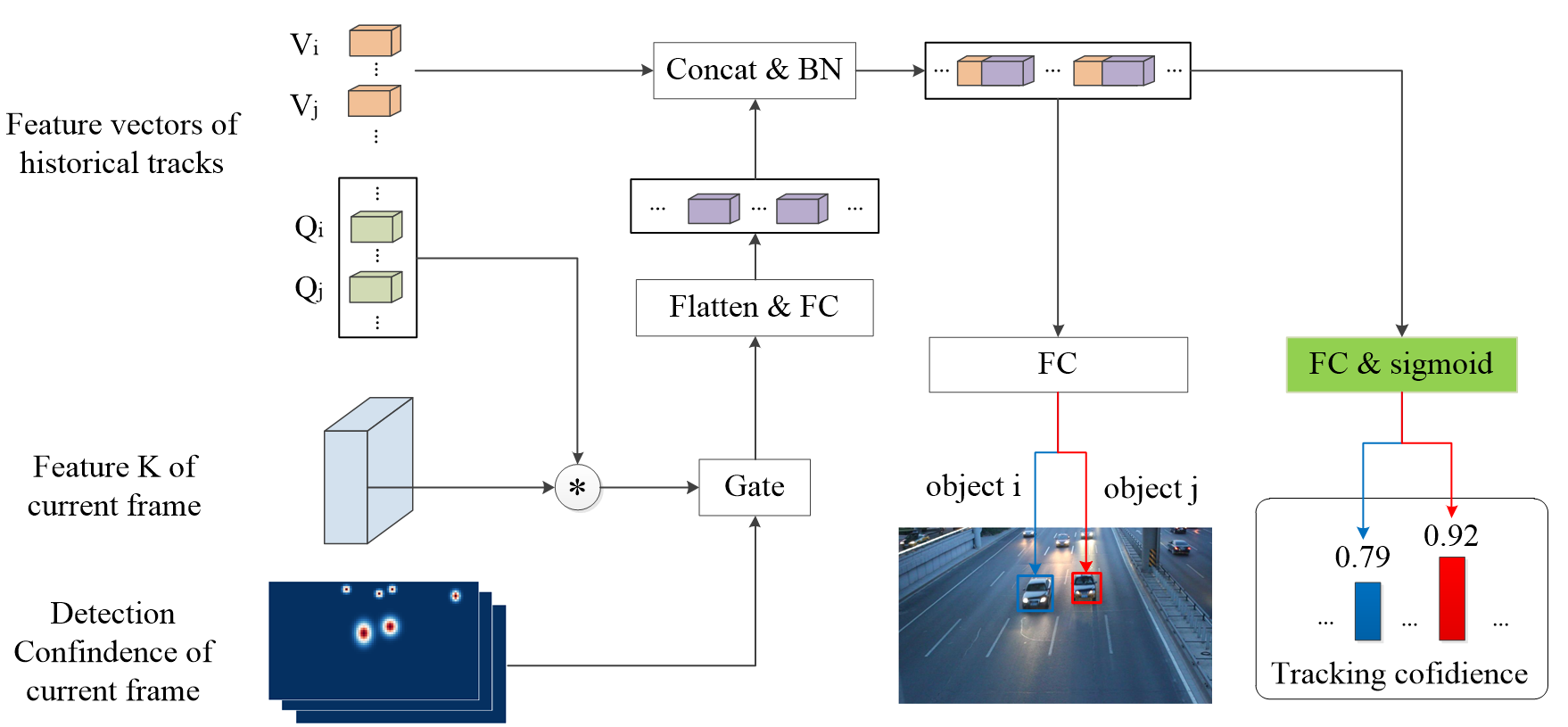}
  \caption{Tracking Module Architecture. This architecture is very similar to the regression branch in the detection module. The green part means that it does not share parameters with the detection module. At training time, the module outputs all potential tracking bounding boxes $B_t\in R^{h'\times w'\times 4}$ and a tracking confidence map $Y_t\in R^{h'\times w'\times 1}$ between frame $t-i$ and $t$, where $t$ is between $1$ and $5$. And only tracking bounding boxes and confidence of historical tracks are calculated at test time.}
  \label{fig:trackingmodule}
\end{figure*}

\textbf{Tracking module}: Tracking is to assign objects in the current frame to historical tracks or generate new tracks. The architecture of the tracking module is depicted as Figure \ref{fig:trackingmodule}. The inputs of the tracking module are: 1) feature map $K$ of the current frame, 2) detection confidence map of the current frame and 3) the feature vectors of historical tracks. And the tracking module outputs a tracking confidence and a bounding box for each historical track. As we can see, this architecture is almost the same as that of the detection module. Most of its network parameters are shared with the detection module, except for the fully connected layer for calculating tracking confidence (the green block in Figure \ref{fig:trackingmodule}). The tracked bounding boxes is consistent with that of the detected target box in the terms of expression, which is $B_i=\left[x_i,y_i,h_i,w_i\right]$, with absolute coordinates and size. The tracking confidences indicate whether the objects are still in the image of the current frame. The tracking module works in an object-wise way, so it can naturally pass the ID of each object to the next frame, which is similar to the parallel single-object tracking (SOT).
\subsection{Training}
Our model can be trained end-to-end, but here, we train the GCNet in two stages. First, train the detection module, and then fine-tune the entire network. The training strategy of the classification branch is consistent with CornerNet. We define a heatmap $Y_{gt}\in R^{h'\times w'\times n}$ with 2D Gaussian kernel as equation (\ref{eq:guasskernel}).
\begin{equation}\label{eq:guasskernel}
\begin{split}
  & Y_{gt,ijk}=\max_{1\leqslant n \leqslant N_k}\left(G_{ijn}\right)  \\
  & G_{ijn}=\exp\left[-\frac{\left(i-x_n\right)^2}{2 \sigma_{x,n}^2 }-\frac{\left(i-y_n\right)^2}{2 \sigma_{y,n}^2 }\right]
\end{split}
\end{equation}
where $N_k$ is the number of objects of class $k$, $\left[x_n,y_n\right]$ is the center of object $n$, and the variance $\sigma^2$ is relative with object size. The expression of $\sigma_{x}$ and $\sigma_{y}$ are equation (\ref{eq:guassvar}), and the $\text{IoU\_threshold}$ is set to $0.3$ here.
\begin{equation}\label{eq:guassvar}
\begin{split}
\sigma_{x}=\frac{h\left(1-\text{IoU\_threshold}\right)}{3\left(1+\text{IoU\_threshold}\right)} \\
\sigma_{y}=\frac{w\left(1-\text{IoU\_threshold}\right)}{3\left(1+\text{IoU\_threshold}\right)} 
\end{split}
\end{equation}
The classification loss is a penalty-reduced pixel-wise focal loss:
{\small
\begin{multline}
  L_{d,cla}=-\frac{1}{h'w'n}\cdot \\
     \sum_{ijk}\begin{cases}
     \left(1-Y_{d,ijk}\right)^2 \log \left(Y_{d,ijk}\right), & Y_{gt,ijk}=1 \\
     \left(1-Y_{gt,ijk}\right)^2 Y_{d,ijk}^{2} \log \left(1-Y_{d,ijk} \right), & Y_{gt,ijk} \neq 1
   \end{cases}
\end{multline}}
The regression branch is trained by CIoU loss, as shown in equation (\ref{eq:lossdetreg}). 
\begin{equation}
L_{d,reg}= \sum_{\left[ij\right]=1} \beta_{ij} \cdot L_{CIoU}\left({B_{gt,ij},B_{d,ij}}\right) \label{eq:lossdetreg}
\end{equation}
where the $[ij]=1$ means the corresponding $B_{d,ij}$ is assigned to a ground truth. We assign a bounding box $B_{d,ij}$ to a ground truth if there are $G_{ijn}>0.3$ and $\sum_n G_{ijn}-\max_n G_{ijn}<0.3$.
{\small
\begin{equation}
[ij]=\begin{cases}
1, \exists_n\! G_{ijn}\!>\!0.3  \& \sum_n\! G_{ijn}\!-\!\max_n\!G_{ijn}\!<\!0.3 \\
0, \qquad\qquad\qquad\qquad \text{otherwise}
\end{cases}
\end{equation}
}

And for $B_{ij}$ with $\max_n G_{ijn}=1$, we set the weights of their regression loss $w_{ij}$  to be $2$, and the weights of others are $1$. This is to enhance the precision of the bounding boxes at center points.

We fine-tune the entire network with a pretrained detection module. At this training step, we input two images $I_{t-i}$ and $I_t$ at once, where $i$ is between $1$ and $5$. The loss contains two parts, detection loss of $I_{t-i}$ and tracking loss between the two images. Tracking loss also consists of two terms, regression CIoU loss and classification focal loss. The tracking ground truth is determined by object ID, the $B_{t,ij}$ and $Y_{t,ij}$ are positive if the $[ij]$ in $I_{t-i}$ equal to $1$ and the corresponding objects also exist in $I_t$. The total train loss is show at equation (\ref{eq:losstotal}).
\begin{equation}
Loss = L_{d,cla} + L_{t,cla} + 0.1 \times(L_{d,reg} + L_{t,reg})  \label{eq:losstotal}
\end{equation}
\subsection{Inference pipeline}
The inference pipeline for joint multi-object detection and tracking is described in Algorithm \ref{alg:inference_pipeline}. The inputs of the algorithm are consecutive frames of images $I_1 \sim I_t$. We record the trajectories $T_i$, confidence $Y_i$ and vectors $[V_i,Q_i]$ of all tracks and candidates in four collections, $\SET{T}$, $\SET{O}$, $\SET{Y}$ and $\SET{C}$, respectively. At each time step, we do object detection on the current frame of image $I$ and track the existing tracks $\SET{T}$ and the candidates $\SET{C}$. We use tracking confidences to update all confidences in sets $\SET{Y}$ and $\SET{C}$, and get $Y_i=\min\left(2\times Y_i \times Y_{t,i},1.5\right)$. The tracks and candidates with a confidence lower than $p_2$ will be deleted, and other trajectories, candidates and corresponding features will be updated. This update strategy, $Y_i=\min\left(2\times Y_i \times Y_{t,i},1.5\right)$, will give those tracks with higher tracking confidence a certain trust margin, and their confidence may be greater than $1$. The detections that have IoU greater than $p_3$ or confidence less than $p_2$  will be ignored. For the remaining detections, those with detection confidence greater than $p_1$ will be used to generated new tracks, and others will be added to the candidate set $\SET{C}$. As we can see, all the detection and tracking can be done in a sparse mode, so the overall computational complexity of the algorithm is very low.
\begin{algorithm}[htbp]
\caption{Inference pipeline of GCNet}
\label{alg:inference_pipeline}
\KwIn{continuous frame images $I_1\sim I_t$.}
\KwOut{object trajectories $\SET{T}=[T_1,T_2,\dots,T_n]$, $T_i=[B_{i,1},B_{i,2},\dots,B_{i,t-1}]$, $B$ denotes bounding box.}
\textbf{Initialize:} trajectory set $\SET{T}=\varnothing$, confidence set $\SET{Y}=\varnothing$, feature vector set $\SET{O}=\varnothing$, candidate set $\SET{C}=\varnothing$ and hyperparameters $p_1$, $p_2$, $p_3$.\\
\For{$I$ \textbf{in} $I_2\sim I_t$}{
  $Q, K, V, B_d = DetectionModule(I)$;\\
  \For{$T_i$ \textbf{in} $\SET{T}$}{
    $B_{t,i}, Y_{t,i} = TrackingModule(Q_i,K,V_i)$;\\
    Update $Y_i=\min\left(2\times Y_i\times Y_{t,i},1.5\right)$;\\
    \eIf{$Y_i < p_2$}{
      Delete $T_i$ from $\SET{T}$;\\
    }{
      Add $B_{t,i}$ to $T_i$;\\
      Update $Q_i=K_{mn}$, $V_i=V_{mn}$,\\
      where $(m,n)$ is the center of $B_{t,i}$;\\
    }
  }
  \For{$C_i=[Y_i,Q_i,V_i,B_i]$ \textbf{in} $\SET{C}$}{
    $B_{t,i}, Y_{t,i} = TrackingModule(Q_i,K,V_i)$;\\
    Update $Y_i=\min\left(2\times Y_i\times Y_{t,i},1.5\right)$;\\
    \eIf{$Y_i < p_1$}{
      Delete $C_i$ from $\SET{C}$;\\
    }{
      Add $B_{t,i}$ to $T_i$;\\
      Update $Q_i=K_{mn}$, $V_i=V_{mn}$,\\
      where $(m,n)$ is the center of $B_{t,i}$;\\
    }
  }
  \For{$B_i$ \textbf{in} $B_d$}{
    \uIf{$\exists\, j , \text{IoU}(B_i,T_{j})>p_3$}{
      \textbf{continue};
    }
    \uElseIf{$Y_i>p_1$}{
      Add $T_{new}=[B_i]$ to $\SET{T}$;\\
      Add $[Q_i,V_i]$ to $\SET{O}$; \\
      Add $Y_i$ to $\SET{Y}$;\\
    }
    \ElseIf{$Y_i>p_2$}{
      Add $[Y_i,Q_i,V_i,B_i]$ to $\SET{C}$;
    }
  }
}
\end{algorithm}
\section{Experiments}

\subsection{Benchmark and implementation details}
We conduct the experiments on the vehicle detection and tracking dataset, UA-DETRAC. This dataset contains $100$ sequences, of which $60$ are used for training and the others are used for testing. The data in the training set and test set come from different traffic scenarios, make the test more difficult. The UA-DETRAC benchmark takes average precision (AP) to rank the performance of detectors, and PR-MOTA, PR-MOTP, PR-MT, PR-ML, PR-IDS, PR-FM, PR-FP, and PR-FN scores for tracking evaluation. We refer to \cite{wen_ua-detrac_2015} for further details on the metrics.

All the experiments are done with TensorFlow 2.0. We train our model by Adam on the complete train dataset of UA-DETRAC. The size of input images is $512\times 896$. We add three commonly used data augmentation methods, random horizontal flip, random brightness adjustment and scale adjustment. The hyperparameters $p_1$, $p_2$, $p_3$ for inference are set to $0.5$, $0.3$ and $0.5$ respectively.

\subsection{Ablation study}
In our join detection and tracking framework, three main tricks may influence the performance: 1) Gate by confidence map $Y_d$, 2) concat the feature vector in $V$ for bounding box regression and 3) specially designed position embedding $P$. We compared the detection effects of the three models with our whole GCNet to show the effectiveness of those tricks. Table \ref{table:ablationstudy} shows the results. This full version of GCNet has the best performance with $74.04\%$ AP on UA-DETRAC. The gate and feature vector of $V$ both yield $2\%$ AP. The Gate step explicitly merges the classification result into the regression branch, which not only plays the role of spatial attention but also conducive to the training of the regression branch. The concated feature vectors of $V$ for regression can introduce more texture and local information, that is not included in correlation vectors, and those information is beneficial to inferring the size of objects. To show the role of position embedding, we replace it with a normal explicit position embedding, where $P_{ijk}$ equals to $i$ when $0\leqslant k<c/2$ and $j$ when $c/2\leqslant k<c$. We can see our self-designed position embedding can take a 5.80\% increase in AP.

\begin{table}[b]
\centering
\caption{Ablation Study results}
\label{table:ablationstudy}
\begin{tabular}{lcccc}
\toprule
\multirow{2}*{Model} & \multicolumn{4}{c}{AP} \\
\cmidrule{2-5}
& Full & Easy & Medium & Hard \\
\midrule
GCNet & \textbf{74.04} & \textbf{91.57} & \textbf{81.45} & \textbf{59.43}\\
Without Gate & 71.62 & 88.49 & 78.99 & 57.56 \\
Without V & 71.71 & 90.29 & 78.13 & 57.65 \\
explicit Pos. Ebd. & 68.24 & 85.28 & 75.59 & 54.61\\ \bottomrule
\end{tabular}
\end{table}

The ablation study is only conducted on the detection benchmark. This is because our tracking module shares most of the parameters with the detection module and tracking performance is highly correlated with the detection performance. Those ablation study results can be extended to the tracking module.
\subsection{Benchmark Evaluation}
\begin{figure*}[hptb]
  \centering
  \includegraphics[scale=0.45]{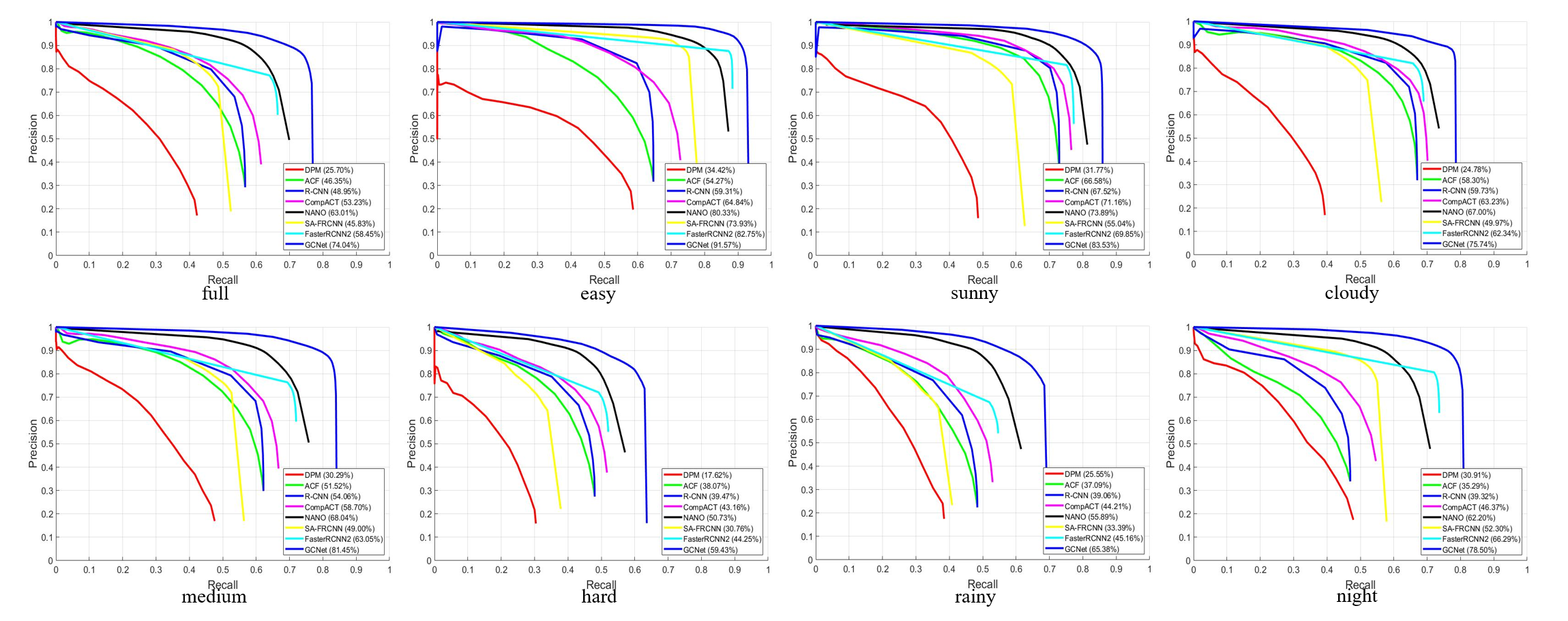}
  \caption{Precision vs. recall curves of the detection algorithms. The scores in the legend are the AP scores for evaluating the performance of object detection algorithms.}
  \label{fig:prcurves}
\end{figure*}

Table \ref{table:detbenchmark} shows the results on UA-DETRAC detection benchmark. Our GCNet demonstrates promising performance and outperforms most detection algorithms on this benchmark. It gets high AP on full, medium difficulty, night, and rainy parts of the test set. Figure \ref{fig:prcurves} shows the PR curves of GCNet and other algorithms exposed by the UA-DETRAC dataset, and we can see our model is far more effective than these baselines in each scenario. It is worth mentioning that our model does not use any other tricks for better precision and the backbone network is only the original version of ResNet50. When only using the detection module of GCNet, it can run at $36$ FPS on a single Nvidia 2080Ti.

\begin{table*}[htbp]
\centering
\caption{Results on UA-DETRAC detection benchmark}
\label{table:detbenchmark}
\begin{tabular}{m{10em}m{4em}m{4em}m{4em}m{4em}m{4em}m{4em}m{4em}}
\toprule
\multirow{2}*{Model} & \multicolumn{7}{c}{AP} \\
\cmidrule{2-8}
& Full & Easy & Medium & Hard & Sunny & Night & Rainy\\
\midrule
DPM & 25.70 & 34.42 & 30.29 & 17.62 & 31.77 & 30.91 & 25.55 \\
ACF & 46.35 & 54.27 & 51.52 & 38.07 & 66.58 & 39.32 & 39.06 \\
R-CNN & 48.95 & 59.31 & 54.06 & 39.47 & 67.52 & 39.32 & 39.06 \\
CompACT & 53.23 & 64.84 & 58.70 & 43.16 & 71.16 & 46.37 & 44.21 \\
Faster R-CNN & 62.13 & 86.14 & 66.77 & 47.29 & 73.83 & 69.28 & 49.03 \\
EB & 67.96 & 89.65 & 73.12 & 53.64 & 83.73 & 73.93 & 53.40 \\
R-FCN & 69.87 & \textbf{93.32} & 75.67 & 54.31 & \textbf{84.08} & 75.09 & 56.21 \\
CenterNet-Res50 & 63.85 & 83.35 & 70.19 & 49.56 & 80.09 & 62.54 & 50.91 \\
GCNet & \textbf{74.04} & 91.57 & \textbf{81.45} & \textbf{59.43} & 83.53 & \textbf{78.50} & \textbf{65.38}\\
\bottomrule
\end{tabular}
\end{table*}

The starting point of designing GCNet is not only for object detection but also for multi-object tracking. This is the real purpose of introducing the global correlation layer to regress the absolute coordinates. And the tracking results are shown in table \ref{table:trackbenchmark}. Those MOT metrics with “PR-” can evaluate the overall effect of detection and tracking. The EB and KIOU is the UA-DETRAC challenge winner. And as we can see, we get a great PR-MOTA score and excellent PR-MOTP score, around twice as high as that of EB+KIOU. Besides, we get the best scores of PR-ML and PR-FN on the UA-DETRAC tracking benchmark. Because the most of features can be shared by the detection and tracking modules, the calculation of the whole joint detection and tracking pipeline is almost the same as that of detection, and it can achieve around 34 FPS.
\begin{table*}[htbp]
\centering
\caption{Results on UA-DETRAC tracking benchmark}
\label{table:trackbenchmark}
\begin{tabular}{lcccccccc}
\toprule
Model & PR-MOTA & PR-MOTP & PR-MT & PR-ML & PR-IDS & PR-FM & PR-FP & PR-FN\\
\midrule
DPM+GOG & 5.5 & 28.2 & 4.1 & 27.7 & 1873.9 & 1988.5 & 38957.6 & 230126.6\\
ACF+GOG & 10.8 & 37.6 & 12.2 & 22.3 & 3850.8 & 3987.3 & 45201.5 & 197094.2\\
R-CNN+DCT & 11.7 & 38.0 & 10.1 & 22.8 & 758.7 & 742.9 & 36561.2 & 210855.6\\
CompACT+TBD & 13.6 & 37.3 & 15.3 & 19.3 & 2026.9 & 2467.3 & 43247.8 & 173837.3\\
CompACT+GOG & 14.2 & 37.0 & 13.9 & 19.9 & 3334.6 & 3172.4 & 32092.9 & 180183.8\\
Faster R-CNN + MHT & 14.5 & 32.5 & 15.9 & 19.1 & 492.3 & \textbf{576.7} & 18141.4 & 156227.8\\
EB+IOU & 19.4 & 28.9 & 17.7 & 18.4 & 2311.3 & 2445.9 & \textbf{14796.5} & 171806.8\\
EB+KIOU & \textbf{21.1} & 28.6 & \textbf{21.9} & 17.6 & \textbf{462.2} & 712.1 & 19046.9 & 159178.3\\
GCNet & 19.1 & \textbf{57.3} & 20.9 & \textbf{9.6} & 755.8 & 994.5 & 17660.9 & \textbf{148517.5}\\
\bottomrule
\end{tabular}
\end{table*}
\balance
\section{Conclusion}
In this paper, we presented a novel joint multi-object detection and tracking network called GCNet. We introduced the global correlation layer to achieve absolute coordinate and size regression, which can not only do object detection on a single image but also naturally propagate the ID of objects to the subsequent consecutive frames. Compared with existing tracking-by-detection methods, our GCNet can achieve end-to-end object trajectories calculation without bounding box NMS, data association, and other complex tracking strategies. We evaluated our method on the UA-DETRAC, a vehicle detection and tracking dataset. And the evaluation results demonstrate our approach has promising effectiveness, not only in detection but also in tracking. Our approach can also run 36 FPS for detection and 34 FPS for joint detection and tracking, which can meet the real-time requirements of most application scenarios, such as autonomous driving and traffic monitoring. 

We hope that this paper can have some enlightenment in the field of joint detection and tracking, and we also believe our approach still has a lot of room for progress and hope it can get sufficient development in future research.
\balance
\nocite{*}
\bibliography{reference}

\end{document}